\documentclass[11pt,a4paper]{article}
\usepackage[hyperref]{emnlp-ijcnlp-2019}
\usepackage{times}
\usepackage{latexsym}
\usepackage{booktabs}
\usepackage{multirow}
\usepackage{url}
\usepackage{amssymb}
\usepackage{pifont}
\usepackage{pgfplots}

\newcommand{\cmark}{\ding{51}}
\newcommand{\xmark}{\ding{55}}

\DeclareSymbolFont{extraup}{U}{zavm}{m}{n}
\DeclareMathSymbol{\varheart}{\mathalpha}{extraup}{86}
\DeclareMathSymbol{\vardiamond}{\mathalpha}{extraup}{87}

\newcommand\independent{\protect\mathpalette{\protect\independenT}{\perp}}
\def\independenT#1#2{\mathrel{\rlap{$#1#2$}\mkern2mu{#1#2}}}

\usepackage{color, colortbl}
\usepackage{amsmath}
\DeclareMathOperator*{\argmax}{arg\,max}

\definecolor{gray}{gray}{0.95}
\definecolor{cyan}{rgb}{0.88,1,1}
\definecolor{lightyellow}{rgb}{1.0, 1.0, 0.88}

\aclfinalcopy 

\title{MRQA 2019 Shared Task: \\ Evaluating Generalization in Reading Comprehension}

\author{Adam Fisch$^\vardiamond$ \hspace{0.15em} Alon Talmor$^{\spadesuit\Diamond}$ \hspace{0.15em}  Robin Jia$^\clubsuit$ \hspace{0.15em} Minjoon Seo$^{\varheart\triangle}$  \hspace{0.15em} Eunsol Choi$^{\varheart\Box}$  \hspace{0.15em}  Danqi Chen$^\heartsuit$\\
$^\vardiamond$ Massachusetts Institute of Technology $^\spadesuit$ Tel Aviv University $^\clubsuit$ Stanford University \\  $^\varheart$ University of Washington  $^\triangle$ NAVER $^\heartsuit$ Princeton University \\ $^\Box$ Google AI $^\Diamond$ Allen Institute for Artificial Intelligence \\
}

\date{}

\begin{document}
\maketitle

\begin{abstract}
We present the results of the Machine Reading for Question Answering (MRQA) 2019 shared task on evaluating the generalization capabilities of reading comprehension systems.\footnote{\href{https://github.com/mrqa/MRQA-Shared-Task-2019}{https://github.com/mrqa/MRQA-Shared-Task-2019}.} In this task, we adapted and unified 18 distinct question answering datasets into the same format. Among them, six datasets were made available for training, six datasets were made available for development, and the final six were hidden for final evaluation. Ten teams submitted systems, which explored various ideas including data sampling, multi-task learning, adversarial training and ensembling.
The best system achieved an average F1 score of 72.5 on the 12 held-out datasets, 10.7 absolute points higher than our initial baseline based on BERT.
\end{abstract}

\section{Introduction}
Machine Reading for Question Answering (MRQA) 
has become an important testbed for evaluating how well computer systems understand human language.
Interest in MRQA settings---in which a system must answer a question by reading one or more context documents---has grown rapidly in recent years, fueled especially by the creation of many large-scale datasets~\cite{squad,triviaqa,naturalquestions}.
MRQA datasets have been used to benchmark progress in general-purpose language understanding \cite{Devlin2018BERTPO,Yang2019XLNetGA}.
Interest in MRQA also stems from their use in industry applications, such as search engines \cite{naturalquestions} and dialogue systems~\cite{coqa,choi2018quac}.


While recent progress on benchmark datasets has been impressive, MRQA systems are still primarily evaluated on in-domain accuracy. 
It remains challenging to build MRQA systems that generalize to new test distributions \cite{chen2017, relationextraction, yogatama2019} and are robust to test-time perturbations \cite{jia2017, ribeiro2018}. A truly effective question answering system should do more than merely interpolate from the training set to answer test examples drawn from the same distribution: it should also be able to extrapolate to test examples drawn from different distributions.

In this work we introduce the MRQA 2019 Shared Task on Generalization, which tests extractive question answering models on their ability to generalize to data distributions different from the distribution on which they were trained. 
Ten teams submitted systems, many of which improved over our provided baseline systems.
The top system, which took advantage of newer pre-trained language models \cite{Yang2019XLNetGA,Zhang2019ERNIEEL},
achieved an average F1 score of 72.5 on our hidden test data, an improvement of 10.7 absolute points over our best baseline.
Other submissions explored using adversarial training, multi-task learning, and better sampling methods to improve performance.
In the following sections, we present our generalization-focused, extractive question-answering dataset, a review of the official baseline and participating shared task submissions, and a meta-analysis of system trends, successes, and failures.

\section{Task Description}
The MRQA 2019 Shared Task focuses on generalization to \emph{out-of-domain} data. 
Participants trained models on a fixed training dataset containing examples from six QA datasets. 
We then evaluated their systems on examples from 12 held-out test datasets. 
For six of the test datasets, we provided participants with some development data;
the other six datasets were entirely hidden---participants did not know the identity of these datasets.

We restricted the shared task to English-language extractive question answering:
systems were given a question and context passage, and were asked to find a segment of text in the context that answers the question.
This format is used by several commonly-used reading comprehension datasets, including SQuAD \cite{squad} and TriviaQA \cite{triviaqa}.
We found that the extractive format is general enough that we could convert many other existing datasets into this format.
The simplicity of this format allowed us to focus on out-of-domain generalization, instead of other important but orthogonal challenges.\footnote{Notatably, the task does \emph{not} test unanswerable \cite{squad_2.0}, multi-turn \cite{coqa}, or open-domain \cite{chen2017} question types.}

The datasets we used in our shared task are given in Table~\ref{tab:splits}. The datasets differ in the following ways:

\begin{itemize}
    \item {\bf Passage distribution:} Context passages come from many different sources, including Wikipedia, news articles, Web snippets, and textbooks. 
    \item {\bf Question distribution:} Questions are of different styles (e.g., entity-centric, relational) and come from different sources, including crowdworkers, domain experts, and exam writers.
    \item {\bf Joint distribution:} The relationship between the passage and question also varies. Some questions were written based on the passage, while other questions were written independently, with context passages retrieved afterwards.
    Some questions were constructed to require multi-hop reasoning on the passage.
\end{itemize}

\paragraph{Evaluation criteria}
Systems are evaluated using exact match score (EM) and word-level F1-score (F1), as is common in extractive question answering tasks \cite{squad, triviaqa, hotpotqa}. EM only gives credit for predictions that exactly match (one of) the gold answer(s), whereas F1 gives a partial credit for partial word overlap with the gold answer(s). We follow the SQuAD evaluation normalization rules and ignore articles and punctuation when computing EM and F1 scores. While more strict evaluation~\cite{naturalquestions} computes scores based on the token indexes of the provided context, we compute scores based on answer string match (i.e., the prediction doesn't need to come from exact same annotated span as long as the predicted answer string matches the annotated answer string). We rank systems based on their macro-averaged test F1 scores across the 12 test datasets.
\begin{table*}
\footnotesize
\centering
\tabcolsep=0.25cm
\begin{tabular}{llllrrcrrr@{}}
\toprule
& \textbf{Dataset}                               & \textbf{Question (Q)} & \textbf{Context (C)} & \textbf{$|$Q$|$} & \textbf{$|$C$|$} &  \textbf{Q $\independent$ C} & \textbf{Train} & \textbf{Dev} & \textbf{Test} \\ \midrule
\multirow{6}{*}{I} & SQuAD                             & Crowdsourced      & Wikipedia  & 11  &  137  & \xmark            & 86,588         & 10,507       & -             \\
& NewsQA                           & Crowdsourced      & News articles    & 8 & 599  & \cmark & 74,160         & 4,212        & -             \\
& TriviaQA$^\spadesuit$                        & Trivia            & Web snippets   & 16 & 784 & \cmark          & 61,688         & 7,785        & -             \\
& SearchQA$^\spadesuit$                         & Jeopardy          & Web snippets & 17 & 749 & \cmark                      & 117,384        & 16,980       & -             \\
& HotpotQA                        & Crowdsourced      & Wikipedia    &  22 & 232 & \xmark                      & 72,928         & 5,904        & -             \\
& Natural Questions        & Search logs       & Wikipedia     & 9  & 153 & \cmark                      & 104,071        & 12,836       & -             \\ \midrule
\multirow{6}{*}{II} & BioASQ$^\spadesuit$                            & Domain experts    & Science articles & 11 & 248  & \cmark                      & -              & 1,504        & 1,518         \\
& DROP$^\diamondsuit$                               & Crowdsourced      & Wikipedia &   11  &  243  & \xmark                      & -              & 1,503        & 1,501         \\
& DuoRC$^\diamondsuit$                              & Crowdsourced      & Movie plots   & 9 & 681 & \cmark                      & -              & 1,501        & 1,503         \\
& RACE$^{\heartsuit}$                                & Domain experts          & Examinations   & 12 & 349 & \xmark                      & -              & 674          & 1,502         \\
& RelationExtraction$^\spadesuit$    & Synthetic         & Wikipedia    & 9 &  30  & \cmark                      & -              & 2,948        & 1,500         \\
& TextbookQA$^{\heartsuit}$                     & Domain experts    & Textbook & 11   &   657  & \xmark                      & -              & 1,503        & 1,508         \\ \midrule
\multirow{6}{*}{III} & BioProcess$^{\heartsuit}$                    & Domain experts    & Textbook   & 9  &  94  & \xmark                      & -              & -            & 219           \\
& ComplexWebQ$^\spadesuit$   & Crowdsourced         & Web snippets   & 14 & 583 & \cmark                      & -              & -            & 1,500         \\
& MCTest$^{\heartsuit}$                           & Crowdsourced      & Crowdsourced  & 9  & 244 & \xmark                      & -              & -            & 1,501         \\
& QAMR$^{\diamondsuit}$                                & Crowdsourced         & Wikipedia  &  7 &  25 & \xmark                      & -              & -            & 1,524         \\
& QAST                                & Domain experts      & Transcriptions  & 10 & 298 & \xmark                      & -              & -            & 220           \\
& TREC$^\spadesuit$         & Crowdsourced      & Wikipedia  & 8  & 792  & \cmark                      & -              & -            & 1,021         \\ \bottomrule
\end{tabular}
\caption{MRQA sub-domain datasets. The first block presents six domains used for training, the second block presents six given domains used for evaluation during model development and the last block presents six hidden domains used for evaluation. $|\cdot|$ denotes the average length in tokens of the quantity of interest. $Q \independent C$ is true if the question was written independently from the passage used for context. 
$\spadesuit$-marked datasets used distant supervision to match questions and contexts,
$\heartsuit$-marked datasets were originally multiple-choice, and
$\diamondsuit$-marked datasets are other datasets where only the answer string is given (rather than the exact answer span in the context).}

\label{tab:splits}
\end{table*}

\section{Dataset Curation}

The MRQA 2019 Shared Task dataset is comprised of many sub-domains, each collected from a separate dataset. The dataset splits and sub-domains are detailed in Table~\ref{tab:splits}. As part of the collection process, we adapted each dataset to conform to the following unified, extractive format:

\begin{enumerate}
    \item The answer to each question must appear as a span of tokens in the passage.
    \item Passages may span multiple paragraphs or documents, but they are concatenated and truncated to the first 800 tokens.
    This eases the computational requirements for processing large documents efficiently.
\end{enumerate}

The first requirement is motivated by the following reasons:

\begin{itemize}
\item Extractive settings are easier to evaluate with stable metrics than abstractive settings.
\item Unanswerable questions are hard to synthesize reliably on datasets without them.
We investigated using distant supervision to automatically generate unanswerable questions, but found it would introduce a significant amount of noise.
\item It is easier to convert multiple-choice datasets to extractive datasets than converting extractive datasets to multiple-choice, as it is difficult to generate challenging alternative answer options.
\item 
Many of popular benchmark datasets are already extractive (or have extractive portions).
\end{itemize} 


\subsection{Sub-domain Splits}
\label{sec:domain-split}
We partition the 18 sub-domains in the MRQA dataset into three splits:

\paragraph{Split I} These sub-domains are available for model training and development, but are not included in evaluation.

\paragraph{Split II} These sub-domains are not available for model training, but are available for model development. Their hidden test portions are included in the final evaluation.

\paragraph{Split III} These sub-domains are not available for model training or development. They are completely hidden to the participants and only used for evaluation.

\paragraph{}Additionally, we balance the testing portions of Splits II and III by re-partitioning the original sub-domain datasets so that we have 1,500 examples per sub-domain. We partition by context, so that no single context is shared across both development and testing portions of either Split II or Split III.\footnote{We draw examples from each dataset's original test split until it is exhausted, and then augment if necessary from the train and dev splits. This preserves the integrity of the original datasets by ensuring that no original test data is leaked into non-hidden splits of the MRQA dataset.}

\subsection{Common Preprocessing}
Datasets may contain contexts that are comprised of multiple documents or paragraphs. We concatenate all documents and paragraphs together. We separate documents with a \texttt{[DOC]} token, insert \texttt{[TLE]} tokens before each document title (if provided), and separate paragraphs within a document with a \texttt{[PAR]} token.

Many of the original datasets do not have labeled answer spans. For these datasets we provide all occurrences of the answer string in the context in the dataset. 
Additionally, several of the original datasets contain multiple-choice questions. For these datasets, we keep the correct answer if it is contained in the context, and discard the other options. We filter questions that depend on the specific options (e.g., questions of the form \emph{``which of the following...''} or \emph{``examples of ... include''}). Removing multiple-choice options might introduce ambiguity (e.g., if multiple correct answers appear in the context but not in the original options). For these datasets, we attempt to control for quality by manually verifying random examples. 




\subsection{Sub-domain Datasets}
In this section we describe the datasets used as sub-domains for MRQA. We focus on the modifications made to convert each dataset to the unified MRQA format. Please see Table~\ref{tab:splits} as well as the associated dataset papers for more details on each sub-domain's properties.

\paragraph{SQuAD \cite{squad}} We used the SQuAD (\textbf{S}tanford \textbf{Qu}estion \textbf{A}nswering \textbf{D}ataset) dataset as the basis for the shared task format.\footnote{A few paragraphs are long, and we discard the QA pairs that do not align with the first 800 tokens (1.1\% of examples).} Crowdworkers are shown paragraphs from Wikipedia and are asked to write questions with extractive answers.

\paragraph{NewsQA \cite{newsqa}} Two sets of crowdworkers ask and answer questions based on CNN news articles. The ``questioners'' see only the article's headline and summary while the ``answerers'' see the full article. We discard questions that have no answer or are flagged in the dataset to be without annotator agreement. 

\paragraph{TriviaQA \cite{triviaqa}} Question and answer pairs are sourced from trivia and quiz-league websites. We use the web version of TriviaQA, where the contexts are retrieved from the results of a Bing search query. 

\paragraph{SearchQA \cite{searchqa}} Question and answer pairs are sourced from the Jeopardy! TV show. The contexts are composed of retrieved snippets from a Google search query. 

\paragraph{HotpotQA \cite{hotpotqa}} Crowdworkers are shown two entity-linked paragraphs from Wikipedia and are asked to write and answer questions that require multi-hop reasoning to solve. In the original setting, these paragraphs are mixed with additional distractor paragraphs to make inference harder. We do not include the distractor paragraphs in our setting.

\paragraph{Natural Questions \cite{naturalquestions}} Questions are collected from information-seeking queries to the Google search engine by real users under natural conditions. Answers to the questions are annotated in a retrieved Wikipedia page by crowdworkers. Two types of annotations are collected: 1) the HTML bounding box containing enough information to completely infer the answer to the question (Long Answer), and 2) the sub-span or sub-spans within the bounding box that comprise the actual answer (Short Answer). 
We use only the examples that have short answers, and use the long answer as the context.

\paragraph{BioASQ \cite{bioasq}} BioASQ, a challenge on large-scale
biomedical semantic indexing and question answering, contains question and answer pairs that are created by domain experts. They are then manually linked to multiple related science (PubMed) articles. We download the full abstract of each of the linked articles to use as individual contexts (e.g., a single question can be linked to multiple, independent articles to create multiple QA-context pairs). We discard abstracts that do not exactly contain the answer.

\paragraph{DROP \cite{drop}} DROP (Discrete
Reasoning Over the content of Paragraphs) examples were collected similarly to SQuAD, where crowdworkers are asked to create question-answer pairs from Wikipedia paragraphs. The questions focus on quantitative reasoning, and the original dataset contains non-extractive numeric answers as well as extractive text answers. We restrict ourselves to the set of questions that are extractive.

\paragraph{DuoRC \cite{duorc}} We use the ParaphraseRC split of the DuoRC dataset. In this setting, two different plot summaries of the same movie are collected---one from Wikipedia and the other from IMDb. Two different sets of crowdworkers ask and answer questions about the movie plot, where the ``questioners'' are shown only the Wikipedia page, and the ``answerers'' are shown only the IMDb page. We discard questions that are marked as unanswerable.

\paragraph{RACE \cite{race}} ReAding Comprehension Dataset From Examinations (RACE) is collected from English reading comprehension exams for middle and high school Chinese students. We use the high school split (which is more challenging) and also filter out the implicit ``fill in the blank'' style questions (which are unnatural for this task).

\paragraph{RelationExtraction \cite{relationextraction}} Given a slot-filling dataset,\footnote{The authors use the WikiReading dataset \cite{wikireading} for the underlying slot-filling task.} relations among entities are systematically transformed into question-answer pairs using templates. For example, the $educated\_at(x, y)$ relationship between two entities $x$ and $y$ appearing in a sentence can be expressed as \emph{``Where was $x$ educated at?''} with answer $y$. Multiple templates for each type of relation are collected. We use the dataset's zero-shot benchmark split (generalization to unseen relations), and only keep the positive examples.  

\paragraph{TextbookQA \cite{textbookqa}} TextbookQA is collected from lessons from middle school Life Science, Earth Science, and Physical Science textbooks. We do not include questions that are accompanied with a diagram, or that are ``True or False'' questions.

\paragraph{BioProcess \cite{bioprocess}} Paragraphs are sourced from a biology textbook, and question and answer pairs about those paragraphs are then created by domain experts.

\paragraph{ComplexWebQ \cite{complexwebquestions}}
ComplexWebQuestions is collected by crowdworkers who are shown compositional, formal queries against Freebase, and are asked to re-phrase them in natural language. Thus, by design, questions require multi-hop reasoning. For the context, we use the default web snippets provided by the authors. We use only single-answer questions of type ``composition'' or ``conjunction''.

\paragraph{MCTest \cite{mctest}} Passages accompanied with questions and answers are written by crowdworkers. The passages are fictional, elementary-level, children's stories.

\paragraph{QAMR \cite{qamr}} To construct the Question-Answer Meaning Representation (QAMR) dataset, crowdworkers are presented with an English sentence along with target non-stopwords from the sentence. They are then asked to create as many question-answer pairs as possible that contain at least one of the target words (and for which the answer is a span of the sentence). These questions combine to cover most of the predicate-argument structures present. We use only the filtered\footnote{The questions that are valid and non-redundant.} subset of the Wikipedia portion of the dataset.

\paragraph{QAST \cite{qast}} We use Task 1 of the Question Answering on Speech Transcriptions (QAST) dataset, where contexts are taken from manual transcripts of spoken lectures on ``speech and language processing.'' Questions about named entities found in the transcriptions are created by English native speakers. Each lecture contains around 1 hour of transcribed text. To reduce the length to meet our second requirement ($\leq$ 800 tokens), for each question we manually selected a sub-section of the lecture that contained the answer span, as well as sufficient surrounding context to answer it.

\paragraph{TREC \cite{yodaqa}} The Text REtrieval Conference (TREC) dataset is curated from the TREC QA tasks \cite{trec} from 1999-2002. The questions are factoid. Accompanying passages are supplied using the Document Retriever from \newcite{chen2017}, if the answer is found within the first 800 tokens of any of the top 5 retrieved Wikipedia documents (we take the highest ranked document if multiple documents meet this requirement).

\section{Baseline Model}
\label{sec:model}

We implemented a simple, multi-task baseline model based on BERT~\cite{Devlin2018BERTPO}, following the MultiQA model \cite{talmor2019}. Our method works as follows:

\paragraph{Modeling} Given a question $q$ consisting of $m$ tokens $\{q_1, \ldots, q_m\}$ and a passage $p$ of $n$ tokens $\{p_1, \ldots, p_n\}$, we first concatenate $q$ and $p$ with special tokens to obtain a joint context $\{\texttt{[CLS]}, q_1, \ldots, q_m, \texttt{[SEP]}, p_1, \ldots, p_n, \texttt{[SEP]}\}$. We then encode the joint context with BERT to obtain contextualized passage representations $\{\mathbf{h}_1, \ldots, \mathbf{h}_n\}$. We train separate MLPs to predict start and end indices independently, and decode the final span using $\argmax_{i,j}\{ p_{start}(i) \times p_{end}(j)\}$.

\paragraph{Preprocessing} Following \newcite{Devlin2018BERTPO}, we create $p$ and $q$ by tokenizing every example using a vocabulary of 30,522 word pieces.
As {BERT} accepts a maximum sequence length of 512, we generate multiple chunks $\{p^{(1)}, \ldots, p^{(k)}\}$ per example by sliding a 512 token window (of the joint context, including $q$) over the entire length of the original passage, with a stride of 128 tokens.

\paragraph{Training}
During training we select only the chunks that contain answers. We maximize the log-likelihood of the first occurrence of the gold answer in each of these chunks, and back-propagate into BERT's parameters (and the MLP parameters). At test time we output the span with the maximal logit across all chunks.

\paragraph{Multi-task Training} We sample up to 75K examples from each training dataset, combine them, and create mixed batches of examples from all of the data. We then follow the same training procedure as before on all the composed training dataset batches.

\section{Shared Task Submissions}
Our shared task lasted for 3 months from May to August in 2019. 
All submissions were handled through the CodaLab platform.\footnote{\href{https://worksheets.codalab.org}{https://worksheets.codalab.org}}
In total, we received submissions from 10 different teams for the final evaluation (Table~\ref{tab:partcipants}). Of these, 6 teams submitted their system description paper. We will describe each of them briefly below. 

\begin{table*}
    \centering
    \begin{tabular}{l|l}\toprule
       \textbf{Model}  &  \textbf{Affliation} \\ \midrule
            D-Net~\cite{64:DNet}       & Baidu Inc. \\
           Delphi~\cite{65:Delphi}       & Apple Inc. \\ 
          \rowcolor{gray} FT\_XLNet & Harbin Institute of Technology \\
            HLTC~\cite{63:HLTC}       &  Hong Kong University of Science \& Technology \\ 
             \rowcolor{gray}  BERT-cased-whole-word & Aristo @ AI2 \\
            CLER~\cite{60:CLER} & Fuji Xerox Co., Ltd. \\
             Adv. Train~\cite{62:Adv}       & 42Maru and Samsung Research \\ 
             \rowcolor{gray}  BERT-Multi-Finetune & Beijing Language and Culture University\\
         PAL IN DOMAIN  & University of California Irvine \\
            HierAtt~\cite{61:HierAtt} & Alexandria University \\ 
             \bottomrule
    \end{tabular}
    \caption{List of participants, ordered by the macro-averaged F1 score on the hidden evaluation set. }
    \label{tab:partcipants}
\end{table*}


\subsection{D-Net~\cite{64:DNet} }
The submission from Baidu adopts multiple pre-trained language models (LMs), including BERT~\cite{Devlin2018BERTPO}, XLNet~\cite{Yang2019XLNetGA}, and ERNIE 2.0~\cite{Zhang2019ERNIEEL}. Unlike other submissions which use only one pre-trained LM, they experiment with 1) training LMs with extra raw text data drawn from science questions and search snippets domains, and 2) multitasking with auxiliary tasks such as natural language inference and paragraph ranking~\cite{Williams2017ABC}. Ultimately, however, the final system is an ensemble of an XLNet-based model and an ERNIE-based model, without auxiliary multitask or augmented LM training. 


\subsection{Delphi~\cite{65:Delphi}}
The submission from Apple investigates the effects of pre-trained language models (BERT vs XLNet), various data sampling strategies, and data augmentation techniques via back-translation. Their final submission uses XLNet~\cite{Yang2019XLNetGA} as the base model, with carefully sampled training instances from negative examples (hence augmenting the model with a no-answer option) and the six training datasets. The final submission does not include data augmentation, as it did not improve performance during development. 


\subsection{HLTC~\cite{63:HLTC}}
The submission from HKUST studies different data-feeding schemes, namely shuffling instances from all datasets versus shuffling dataset-ordering only. Their submission is built on top of XLNet, with a multilayer perceptron layer for span prediction. They also attempted to substitute the MLP layer with a more complex attention-over-attention (AoA) \cite{cui2017} layer on top of XLNet, but did not find it to be helpful. 


\subsection{CLER~\cite{60:CLER}}
The submission from Fuji Xerox adds a mixture-of-experts (MoE)~\cite{Jacobs1991AdaptiveMO} layer on top of a BERT-based architecture. They also use a multi-task learning framework trained together with natural language inference (NLI) tasks. Their final submission is an ensemble of three models trained with different random seeds.

\subsection{Adv. Train~\cite{62:Adv}}
The submission from 42Maru and Samsung Research proposes an adversarial training framework, where a domain discriminator predicts the underlying domain label from the QA model's hidden representations, while the QA model tries to learn to arrange its hidden representations such that the discriminator is thwarted. Through this process, they aim to learn domain (dataset) invariant features that can generalize to unseen domains. The submission is built based on the provided BERT baselines.

\subsection{HierAtt~\cite{61:HierAtt}}
The submission from Alexandria University uses the BERT-Base model to provide feature representations. Unlike other models which allowed fine-tuning of the language model parameters during training, this submission only trains model parameters associated with the question answering task, while keeping language model parameters frozen. The model consists of two attention mechanisms: one bidirectional attention layer used to model the interaction between the passage and the question, and one self-attention layer applied to both the question and the passage.

\begin{table*}
\centering
\begin{center}
\begin{tabular}{l|r r |  r r | r}
\toprule
 \textbf{Model} & \textbf{Split I} & \textbf{Split II}  & \textbf{Split II}  &  \textbf{Split III}  & \textbf{Split II + III} \\
\midrule
Portion (\# datasets) & Dev (6) & Dev (6) & Test (6) & Test (6) & Test (12) \\
\midrule
D-Net~\cite{64:DNet} &\textbf{84.1} &  \textbf{69.7} & \textbf{68.9} &\textbf{76.1} &	\textbf{72.5}	\\
Delphi~\cite{65:Delphi} &	82.3	& 68.5	 & 66.9 &	74.6 &70.8 \\
\rowcolor{gray}FT\_XLNet &	82.9 & 68.0	& 66.7 &	74.4 &	70.5 \\
HLTC~\cite{63:HLTC} 	&81.0	& 65.9  & 	65.0&	72.9 &69.0 \\
\rowcolor{gray}BERT-cased-whole-word	& 79.4 &	61.1 &	61.4&	71.2 & 66.3 \\
CLER~\cite{60:CLER}	&80.2	& 62.7 &	62.5&	69.7 &	66.1\\
Adv. Train~\cite{62:Adv} &76.8 & 57.1 &	57.9 &	66.5 	&	62.2 \\
\rowcolor{lightyellow}
Ours: BERT-Large &76.3	 &	57.1 & 57.4&	66.1 	&	61.8\\
\rowcolor{gray}BERT-Multi-Finetune &74.2	 & 53.3	& 56.0 &	64.7 	&	60.3\\
\rowcolor{lightyellow}
Ours: BERT-Base & 74.7& 54.6	&54.6 	&62.4&	58.5 \\
HierAtt~\cite{61:HierAtt} 	&71.1	& 48.7 &	50.5 &	61.7 &56.1 \\
\bottomrule
\end{tabular}
\end{center}
\caption{Performance as F1 score on the shared task. Each score is macro-averaged across individual datasets. The last column (test portion of Split II and III) is used for the final ranking. Our baselines are shaded in yellow, and the submissions which did not present system description papers are shaded in grey. }
\label{tab:main_result}
\end{table*}

\section{Results}
\subsection{Main Results}
\label{sec:main_result}

Table~\ref{tab:main_result} lists the macro-averaged F1 scores of all the submissions on both the development and testing portions of the MRQA dataset. The teams are ranked by the F1 scores on the hidden testing portions of the 12 datasets (Split II and III in Section~\ref{sec:domain-split}). As seen in Table~\ref{tab:main_result}, many of the submissions outperform our BERT-Large baseline significantly. The best-performing system, D-Net~\cite{64:DNet}, achieves an F1 score of 72.5, which is a 10.7 point absolute improvement over our baseline, and 11.5 and 10.0 point improvements, respectively, on Split II (with the development portions provided) and Split III datasets (completely hidden to the participants). 

We evaluate all the submissions on the in-domain datasets (Split I) in Table~\ref{tab:main_result} and find that there is a very strong correlation between in-domain and out-of-domain performance. The top submissions on the out-of-domain datasets also obtain the highest scores on the six datasets that we provided for training.

We present per-dataset performances for 12 evaluation datasets in the appendix. Across the board, many submitted systems greatly outperform our baselines. Among the 12 datasets, performance on the DROP dataset has improved the most---from 43.5 F1 to 61.5 F1---while performance on the RelationExtraction dataset has improved the least (84.9 F1 vs. 89.0 F1). The models with higher average scores seemed to outperform in most datasets: the performance rankings of submissions are mostly preserved on individual datasets.

\begin{table}
\footnotesize
    \centering
    \begin{tabular}{l|l|c|rrr}
\toprule
     \multicolumn{2}{c|}{} & \multirow{2}{*}{\#} & \multirow{2}{*}{Best} & \multirow{2}{*}{Base}  &\multirow{2}{*} {Impr.} \\ 
      \multicolumn{2}{c|}{} & &  &    &  \\ \midrule
  \multirow{3}{*}{\begin{tabular}[c]{@{}c@{}}Question\\ Type\end{tabular}}   & Crowdsourced & 6   &  69.9 & 58.5 & 11.5 \\
  & Synthetic & 1  &  88.9 & 84.7 & 4.2 \\ 
  & Domain experts & 5 & 71.5 & 60.5 & 11.5 \\
  \midrule
\multirow{3}{*}{\begin{tabular}[c]{@{}c@{}}Context\\ Type\end{tabular}}    & Wikipedia &4& 73.4 & 62.3 & 11.1\\
& Education& 4 & 68.2 & 56.2 & 12.0\\
& Others& 4 & 76.1 & 66.8 & 9.3\\
\midrule
\multirow{2}{*}{\textbf{Q $\independent$ C}} & \cmark &5& 73.0 & 63.8 & 9.2 \\
& \xmark &7& 72.2 & 60.3 & 11.9 \\
\bottomrule
    \end{tabular}
    \caption{Macro-averaged F1 scores based on the dataset characteristics as defined in Table~\ref{tab:splits}. {Best} denotes the best submitted system (D-Net), and {Base} denotes our BERT-Large baseline.}
    \label{tab:per_data_type}
\end{table}
\subsection{Summary of Findings}

\paragraph{Improvements per data types}
We analyzed the average performance across the various types of datasets that are represented in Table~\ref{tab:splits}. Table~\ref{tab:per_data_type} summarizes our observations: (1) the datasets with naturally collected questions (either crowdsourced or curated by domain experts) all obtain large improvements; (2) The datasets collected from Wikipedia or education materials (textbooks and Science articles) receive bigger gains compared to those collected from Web snippets or transcriptions; and (3) There is a bigger improvement for datasets in which questions are posed dependent on the passages compared to those with independently collected questions (11.9 vs. 9.2 points).

\paragraph{Pre-trained language models}

The choice of pre-trained language model has a significant impact on the QA performance, as well as the generalization ability. Table~\ref{tab:pretraining} summarizes the pre-trained models each submission is based on, along with its evaluation F1 score. The top three performing systems all use XLNet instead of BERT-Large---this isolated change in pre-trained language model alone yields a significant gain in overall in- and out-of-domain performance. \citet{64:DNet} argues that XLNet shows superior performances on datasets with discrete reasoning, such as DROP and RACE. \citet{63:HLTC}, however, also use XLNet, but does not show strong gains on the DROP or RACE datasets. 

The winning system ensembled two \textit{different} pre-trained language models. Only one other submission~\cite{60:CLER} used an ensemble for their final submission, merging the same LM with different random seeds.

\begin{table}[!h]
    \setlength\extrarowheight{1pt}
    \centering
    \small
    \begin{tabular}{l|c|c}
    \toprule
    \multirow{2}{*}{\textbf{Model}} & \textbf{Base}  & \textbf{Eval F1} \\
    & \textbf{Language Model} & \textbf{(II + III)}\\ \midrule
        D-Net &  XLNet-L + ERNIE 2.0 & 72.5\\
        Delphi & XLNet-L & 70.8\\
        HLTC & XLNet-L & 69.0\\ \midrule
        CLER & BERT-L & 66.1\\
        Adv. Train & BERT-L & 62.2\\
        \rowcolor{lightyellow} BERT-Large & BERT-L	&	61.8\\
        HierAtt & BERT-B & 56.1\\
        \bottomrule
    \end{tabular}
    \caption{Pretrained language models used in the shared task submissions. *-L and *-B denote large and base versions of the models. }
    \label{tab:pretraining}
\end{table}

\pgfplotstableread[row sep=\\,col sep=&]{
    interval & Best & Baseline & Finetune \\
    BioASQ     & 75.3  & 66.6  & 68.2  \\
    DROP     & 61.5 & 43.5  & 52.1  \\
    DuoRC    & 66.6 & 55.1 & 57.5 \\
    RACE   & 53.9 & 41.4 & 45.3 \\
    RelExt   & 89.0  & 84.7 & 90.6 \\
    TextbookQA      & 67.6  & 53.2 & 58.0 \\
    Macro Avg.     & 68.9 & 57.4 & 62.0 \\
    }\mydata
 
\begin{figure*}[!ht]
\small
\centering
\begin{tikzpicture}
    \begin{axis}[
            ybar,
            bar width=.5cm,
            width=\textwidth,
            height=.4\textwidth,
            legend style={column sep=0.05cm, at={(0.3,1)},
                anchor=north,legend columns=-1},
            symbolic x coords={BioASQ,DROP,DuoRC,RACE,RelExt,TextbookQA, Macro Avg.},
            xtick=data,
            ybar=2.2pt,
            nodes near coords,
            nodes near coords align={vertical},
            nodes near coords style={font=\tiny},
            ymin=30,ymax=100,
            ylabel={F1 Score},
        ]
        \addplot table[x=interval,y=Best]{\mydata};
        \addplot table[x=interval,y=Baseline]{\mydata};
        \addplot table[x=interval,y=Finetune]{\mydata};
        \legend{Best result, Baseline, Fine-tuned baseline}
    \end{axis}
\end{tikzpicture}
\caption{F1 scores on Split II sub-domains (test portions) comparing the best submitted system (D-Net) against our BERT-Large baseline. The third result for each dataset is from individually fine-tuning the BERT-Large baseline on the in-domain dev portion of the same dataset (i.e., Split II (dev)).}
\label{fig:finetune}
\end{figure*}
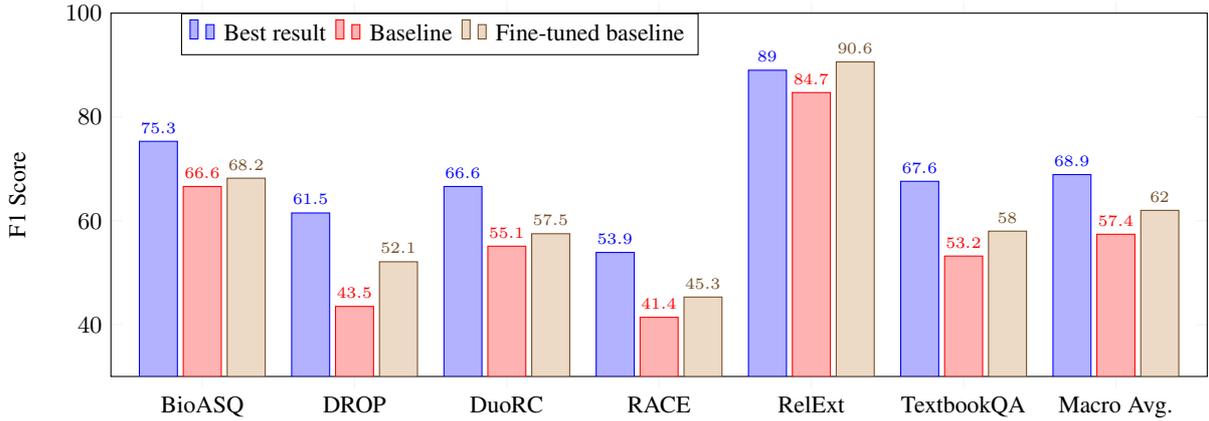

\paragraph{Data sampling}
Our shared task required all participants to use our provided training data, compiled from six question answering datasets, and disallowed the use of any other question-answering data for training.
Within these restrictions, we encouraged participants to explore \textit{how} to best utilize the provided data. 

Inspired by \citet{talmor2019}, two submissions~\cite{63:HLTC,65:Delphi} analyzed similarities between datasets. Unsurprisingly, the performance improved significantly when fine-tuned on the training dataset most similar to the evaluation dataset of interest. \citet{63:HLTC} found each of the development (Split II) datasets resembles one or two training datasets (Split I)---and thus training with all datasets is crucial for generalization across the multiple domains. They experimented with data-feeding methodologies, and found that shuffling instances of all six training datasets is more effective than sequentially feeding all examples from each dataset, one dataset after another.

Additionally, \citet{65:Delphi} observed that the models fine-tuned on SearchQA and TriviaQA achieve relatively poor results across all the evaluation sets (they are both trivia-based, distantly supervised, and long-context datasets). Downsampling examples from these datasets increases the overall performance. They also found that, although our shared task focuses on answerable questions, sampling negative examples leads to significant improvements (up to +1.5 F1 on Split II and up to +4 F1 on Split I). Since most systems follow our baseline model (Section~\ref{sec:model}) by doing inference over \emph{chunks} of tokens, not all examples fed to these models are actually guaranteed to contain an answer span.




\paragraph{Multi-task learning}
Two submissions attempted to learn the question answering model together with other auxiliary tasks, namely natural language inference~\cite{60:CLER,64:DNet} or paragraph ranking~\cite{64:DNet} (i.e., classifying whether given passages contains an answer to the question or not). This could improve the generalization performance on question answering for two reasons. First, the additional training simply exposes the model to more diverse domains, as the entailment dataset~\cite{Williams2017ABC} contains multiple domains ranging from fiction to telephone conversations. Second, reasoning about textual entailment is often necessary for question answering, while passage ranking (or classification) is an easier version of extractive question answering, where the model has to identify the passage containing the answer instead of exact span. 

Both systems introduced task-specific fully connected layers while sharing lower level representations across different tasks. While \citet{60:CLER} showed a modest gain by multi-tasking with NLI tasks (+0.7 F1 score on the development portion of Split II), \citet{64:DNet} reported that multitasking did not improve the performance of their best model. 

\paragraph{Adversarial Training}
One submission~\cite{62:Adv} introduced an adversarial training framework for generalization. The goal is to learn domain-invariant features (i.e., features that can generalize to unseen test domains) by jointly training with a domain discriminator, which predicts the dataset (domain) for each example. 
According to \newcite{62:Adv}, this adversarial training helped on most of the datasets (9 out of 12), but also hurt performance on some of them. It finally led to +1.9 F1 gain over their BERT-Base baseline, although the gain was smaller (+0.4 F1) for their stronger BERT-Large baseline.


\paragraph{Ensembles}
Most extractive QA models, which output a logit for the start index and another for the end index, can be ensembled by adding the start and end logits from models trained with different random seeds. This has shown to improve performances across many model classes, as can be seen from most dataset leaderboards. 
The results from the shared task also show similar trends. A few submissions~\cite{60:CLER,64:DNet} tried ensembling, and all reported modest gains. While ensembling is a quick recipe for a small gain in performance, it also comes at the cost of computational efficiency---both at training and at inference time. 

Related to ensembling, \citet{60:CLER} uses a mixture of experts~\cite{Jacobs1991AdaptiveMO} layer, which learns a gating function to ensemble different weights, adaptively based on the input.




\subsection{Comparison to In-domain Fine-tuning}
Lastly, we report how the best shared task performance compares to in-domain fine-tuning performance of our baseline. Section~\ref{sec:main_result} shows large improvements by the top shared task model, D-Net, over our baseline. We analyze to what extent the reduced performance on out-of-domain datasets can be overcome by exposing the baseline to only a few samples from the target distributions. 
As suggested by \newcite{liu-schwartz-smith:2019:NAACL}, if the model can generalize with a few examples from the new domain, poor performance on that domain is an indicator of a lack of training data diversity, rather than of fundamental model generalization weaknesses. 

Figure~\ref{fig:finetune} presents our results on the six datasets from Split II, where we have individually fine-tuned the BERT-Large baseline on each of the Split II dev datasets and tested on the Split II test datasets. We see that while the gap to D-Net shrinks on all datasets (overall performance increases by 4.6 F1), surprisingly it is only completely bridged in one of the settings (RelationExtraction). This is potentially because this dataset covers only a limited number of relations, so having in-domain data helps significantly.  This suggests that D-Net (and the other models close to it in performance) is an overall stronger model---a conclusion also supported by its gain on in-domain data (Split I).

\section{Conclusions}
We have presented the MRQA 2019 Shared Task, which focused on testing whether reading comprehension systems can generalize to examples outside of their training domain.
Many submissions improved significantly over our baseline, and investigated a wide range of techniques.

Going forward, we believe it will become increasingly important to build NLP systems that generalize across domains.
As NLP models become more widely deployed, they must be able to handle diverse inputs, many of which may differ from those seen during training.
By running this shared task and releasing our shared task datasets,
we hope to shed more light how to build NLP systems that generalize beyond their training distribution.

\section*{Acknowledgements}
We would like to thank Jonathan Berant, Percy Liang, and Luke Zettlemoyer for serving as our steering committee. We are grateful to Baidu, Facebook, and Naver for providing funding for our workshop. We thank Anastasios Nentidis and the entire BioASQ organizing committee for letting us use BioASQ shared task data for our task, and for hosting the data files. We also thank Valerie Mapelli and ELRA for providing us with the QAST data: CLEF QAST (2007-2009) -- Evaluation Package, ELRA catalogue (http://catalog.elra.info), CLEF QAST (2007-2009) -- Evaluation Package, ISLRN: 460-370-870-489-0, ELRA ID: ELRA-E0039.
Finally, we thank the CodaLab Worksheets team for their help with running the shared task submissions.

 \bibliography{emnlp-ijcnlp-2019}
 \bibliographystyle{acl_natbib}
 
\newpage
\section*{Appendix}
We present the per-dataset performances in Table~\ref{tab:result_split2} and Table~\ref{tab:result_split3} for shared task submissions and our baselines. 

\begin{table*}[!ht]
\centering
\footnotesize
\setlength\extrarowheight{1pt}
\begin{tabular}{lcccccccccccc}
\toprule
& \multicolumn{2}{c}{\textbf{BioASQ}} & \multicolumn{2}{c}{\textbf{DROP}} & \multicolumn{2}{c}{\textbf{DuoRC}} & \multicolumn{2}{c}{\textbf{RACE}} & \multicolumn{2}{c}{\textbf{RelExt}} & \multicolumn{2}{c}{\textbf{TextbookQA}} \\ 
\textbf{Model}                          & \textbf{EM}           & \textbf{F1}          & \textbf{EM}          & \textbf{F1}         & \textbf{EM}          & \textbf{F1}          & \textbf{EM}          & \textbf{F1}         & \textbf{EM}               & \textbf{F1}               & \textbf{EM}             & \textbf{F1}            \\ \midrule
D-Net   &  \textbf{61.2} &\textbf{75.3} &	\textbf{50.7} &\textbf{61.5}	& \textbf{54.7} & \textbf{66.6}&	\textbf{39.9}&53.5 &	\textbf{80.1}& \textbf{89.0} &	\textbf{57.2}& 
\textbf{67.6}   \\     
Delphi                         & 60.3        & 72.0       & 48.5       & 58.9      & 53.3       & 63.4       & {39.4}       & \textbf{53.9}      & 79.2            & 87.9            & 56.5          & 65.5         \\
FT\_XLNet                      & 59.3        & 72.9       & 48.0       & 58.3      & 52.7       & 63.8       & 39.4       & 53.8      & 79.0            & 87.2            & 53.6          & 64.2         \\
HLTC                      & 59.6        & 74.0       & 41.0       & 51.1      & 51.7       & 63.1       & 37.2       & 50.5      & 76.5            & 86.2            & 55.5          & 65.2         \\
BERT-cased-whole-word & 57.8        & 72.9       & 43.1       & 53.2      & 42.3       & 53.5       & 35.0       & 48.7      & 78.5            & 87.9            & 43.9          & 51.9         \\
CLER                           & 53.2        & 68.8       & 37.7       & 47.5      & 51.6       & 62.9       & 31.9       & 45.0      & 78.6            & 87.7            & 53.5          & 62.9         \\
Adv. Train           & 45.1        & 60.5       & 34.8       & 43.8      & 46.2       & 57.3       & 29.6       & 42.8      & 74.3            & 84.9            & 48.8          & 58.0         \\
\rowcolor{lightyellow}Ours: BERT-Large                     & 49.7        & 66.6       & 33.9       & 43.5      & 43.4       & 55.1       & 29.0       & 41.4      & 72.5            & 84.7            & 45.6          & 53.2         \\
BERT-Multi-Finetune            & 48.7        & 64.8       & 30.4       & 40.3      & 43.7       & 54.7       & 26.4       & 38.7      & 75.3            & 85.0            & 44.0          & 52.4         \\
\rowcolor{lightyellow}Ours: BERT-Base                      & 46.4        & 60.8       & 28.3       & 37.9      & 42.8       & 53.3       & 28.2       & 39.5      & 73.3            & 83.9            & 44.3          & 52.0         \\
HierAtt               & 43.0        & 59.1       & 24.4       & 34.8      & 38.5       & 49.6       & 24.6       & 37.4      & 67.9            & 81.3            & 32.1          & 40.5         \\ \bottomrule
\end{tabular}
\caption{Performance on the six datasets of Split II (test portion). EM: exact  match, F1: word-level  F1-score. }
\label{tab:result_split2}
\end{table*}

\begin{table*}[!ht]
\small
\setlength\extrarowheight{1pt}
\centering
\begin{tabular}{@{}lcccccccccccc@{}}
\toprule
                                 & \multicolumn{2}{c}{\textbf{BioProcess}} & \multicolumn{2}{c}{\textbf{ComWebQ}} & \multicolumn{2}{c}{\textbf{MCTest}} & \multicolumn{2}{c}{\textbf{QAMR}} & \multicolumn{2}{c}{\textbf{QAST}} & \multicolumn{2}{c}{\textbf{TREC}} \\ 
\textbf{Model}                            & \textbf{EM}             & \textbf{F1}            & \textbf{EM}             & \textbf{F1}             & \textbf{EM}           & \textbf{F1}          & \textbf{EM}          & \textbf{F1}         & \textbf{EM}          & \textbf{F1}         & \textbf{EM}          & \textbf{F1}         \\\midrule
D-NET         &  \textbf{61.3} & \textbf{75.6}&	\textbf{67.8}& \textbf{68.3}& 	67.8& \textbf{80.8}&	60.4 & \textbf{76.1} &75.0 & 	88.8& 51.8&	\textbf{66.8}   \\
Delphi                           & 58.9          & 74.2         & 55.1          & 62.3          & \textbf{68.0}        & 80.2       &\textbf{ 61.0}       & 75.3      & \textbf{78.6}       & \textbf{89.9}      & \textbf{55.0}       & 65.8      \\
FT\_XLNet                        & 62.6          & 75.2         & 54.8          & 62.7          & 66.0        & 79.6       & 56.5       & 73.4      & 76.8       & 90.0      & 51.8       & 65.5      \\
HLTC                       & 56.2          & 72.9         & 54.7          & 61.4          & 64.6        & 78.7       & 56.4       & 72.5      & 75.9       & 88.8      & 49.9       & 63.4      \\
BERT-cased-whole-word   & 56.2          & 71.5         & 52.4          & 60.7          & 63.8        & 76.4       & 56.1       & 71.5      & 69.6       & 85.3      & 43.6       & 61.6      \\
CLER                             & 48.0          & 68.4         & 52.6          & 61.2          & 59.9        & 73.1       & 54.3       & 71.4      & 65.0       & 84.3      & 42.7       & 60.0      \\
Adv. Train             & 46.1          & 62.9         & 48.7          & 56.9          & 57.2        & 70.9       & 56.8       & 71.7      & 56.8       & 77.8      & 42.6       & 58.8      \\
\rowcolor{lightyellow}Ours: BERT-Large                       & 46.1          & 63.6         & 51.8          & 59.1          & 59.5        & 72.2       & 48.2       & 67.4      & 62.3       & 80.8      & 36.3       & 53.6      \\
BERT-Multi-Finetune              & 43.4          & 58.8         & 49.6          & 57.7          & 59.2        & 72.2       & 48.6       & 67.0      & 60.0       & 80.1      & 34.6       & 52.3      \\
\rowcolor{lightyellow}Ours: BERT-Base                        & 38.4          & 57.4         & 47.4          & 55.3          & 54.2        & 66.1       & 47.8       & 64.8      & 58.6       & 77.0      & 36.7       & 54.0      \\
HierAtt  & 44.3          & 60.8         & 41.9          & 51.2          & 54.2        & 67.9       & 48.0       & 66.0      & 50.9       & 75.5      & 27.7       & 48.7      \\ \bottomrule
\end{tabular}
\caption{Results on the six datasets of Split III. EM: exact  match, F1: word-level  F1-score.}
\label{tab:result_split3}
\end{table*}

\end{document}